\documentclass[lettersize,journal]{IEEEtran}
\usepackage{amsmath,amsfonts}
\usepackage{algorithmic}
\usepackage{algorithm}
\usepackage{array}
\usepackage[caption=false,font=normalsize,labelfont=sf,textfont=sf]{subfig}
\usepackage{textcomp}
\usepackage{stfloats}
\usepackage{url}
\usepackage{verbatim}
\usepackage{graphicx}
\usepackage{cite}
\usepackage{booktabs}
\usepackage{color}
\usepackage{graphicx}

\def\blue#1{\textcolor[rgb]{0,0,1}{#1}}

\usepackage{amsmath}
\usepackage{amsfonts}
\usepackage{multirow}
\usepackage{xcolor}
\usepackage{colortbl}
\usepackage{array}
\usepackage{tabularx}

\begin{document}

\title{Domain-invariant Progressive Knowledge Distillation for UAV-based Object Detection}
% Progressive and Domain-invariant Teaching: Excavating the Potential of Student Model in UAV-based Object Detection 
\author{Liang Yao, \IEEEmembership{Graduate Student Member, IEEE}, Fan Liu,  \IEEEmembership{Member, IEEE}, \\ Chuanyi Zhang, \IEEEmembership{Member, IEEE}, Zhiquan Ou, and Ting Wu 
\thanks{This work was partially supported by the Fundamental Research Funds for the Central Universities (No. B240201077), National Nature Science Foundation of China (No. 62372155 and No. 62302149), Aeronautical Science Fund (No. 2022Z071108001), Joint Fund of Ministry of Education for Equipment Pre-research (No. 8091B022123), Water Science and Technology Project of Jiangsu Province under grant No. 2021063, Qinglan Project of Jiangsu Province, Changzhou science and technology project No. 20231313. The work of Liang Yao was supported in part by Postgraduate Research \& Practice Innovation Program of Jiangsu Province (No. SJCX24\_0183). }
\thanks{Liang Yao, Fan Liu, Chuanyi Zhang, Zhiquan Ou, and Ting Wu are with the College of Computer Science and Software Engineering, Hohai University, Nanjing, 210098, China. Corresponding author: Fan Liu (fanliu@hhu.edu.cn).}
}

\markboth{Journal of \LaTeX\ Class Files,~Vol.~14, No.~8, August~2021}%
{Shell \MakeLowercase{\textit{et al.}}: A Sample Article Using IEEEtran.cls for IEEE Journals}


\maketitle

\begin{abstract}
Knowledge distillation (KD) is an effective method for compressing models in object detection tasks. Due to limited computational capability, UAV-based object detection (UAV-OD) widely adopt the KD technique to obtain lightweight detectors. Existing methods often overlook the significant differences in feature space caused by the large gap in scale between the teacher and student models. This limitation hampers the efficiency of knowledge transfer during the distillation process. Furthermore, the complex backgrounds in UAV images make it challenging for the student model to efficiently learn the object features. In this paper, we propose a novel knowledge distillation framework for UAV-OD. Specifically, a progressive distillation approach is designed to alleviate the feature gap between teacher and student models. Then a new feature alignment method is provided to extract object-related features for enhancing student model's knowledge reception efficiency. Finally, extensive experiments are conducted to validate the effectiveness of our proposed approach. The results demonstrate that our proposed method achieves state-of-the-art (SoTA) performance in two UAV-OD datasets. %Our code will be publicly available once the paper is accepted.
\end{abstract}

\begin{IEEEkeywords}
Deep Learning, Knowledge Distillation, UAV-based Object Detection, Fast Fourier Transform, Domain-invariant Feature Learning
\end{IEEEkeywords}

\section{Introduction}

\begin{figure}[t]
    \centering
    \includegraphics[width=0.99\linewidth]{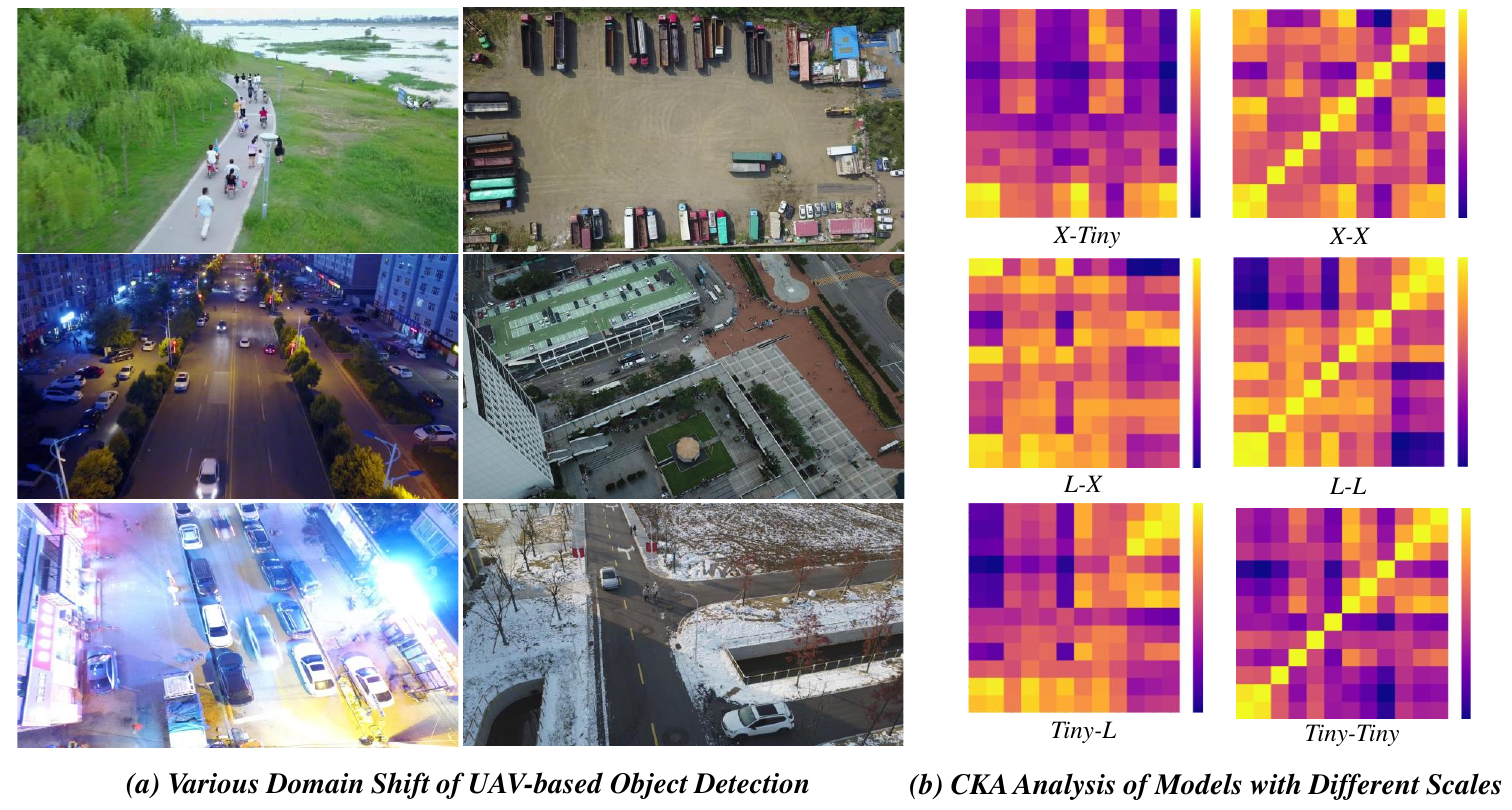}
    % \vspace{-0.3cm}
    \caption{Motivations of our proposed method. \textbf{(a)} The complex environments of UAVs lead to significant domain variations. For instance, factors such as background, lighting, perspective, and weather conditions contribute to the difficulty of UAV-OD. \textbf{(b)} We conducted CKA analysis on features of models at different scales (YOLOv7-X, L, Tiny). The results indicate that closer-scale model features have smaller differences, which are more conducive to knowledge transfer. }
    \label{fig1}
    % \vspace{-0.3cm}
\end{figure}

Unmanned Aerial Vehicles (UAV) equipped with cameras have been exploited in a wide variety of multi-media applications~\cite{huang2022object, luo2023evolutionary, han2022comprehensive, lopez2022unmanned}. As one of the fundamental tasks, UAV-based object detection (UAV-OD) has garnered considerable interest~\cite{zhang2024empowering,wu2021deep,zitar2023review}. However, UAV-OD still remains a challenging task. 
On the one hand, UAVs tend to have limited computational capability, raising an urgent demand in lightweight deployed models for fast inference and low latency. On the other hand, as illustrated in Fig~\ref{fig1} (a), images captured by UAVs exhibit significant variations in background, lighting, and weather conditions. These factors create a large domain shift that complicates accurate object detection.

% On the other hand, the diverse flying environments results in images covering a wide domain gap, making it difficult to accurately detect objects.}

In order to deal with the dilemma of balancing the accuracy and efficiency, several techniques have been proposed, including pruning~\cite{han2015learning,fang2023depgraph}, quantization~\cite{han2015deep}, and knowledge distillation~\cite{hinton2015distilling,li2023object}. Knowledge distillation, due to its simple and effective characteristics, is widely employed for lightweight object detection models~\cite{cao2022pkd,yang2022focal,zhang2023structured,wang2024crosskd}. Existing knowledge distillation methods mainly focus on aligning the intermediate representations (feature-level) of student and teacher models. However, they often directly utilize complex models as teachers and simpler models as students, overlooking the knowledge space differences caused by scale disparities. As illustrated in Fig~\ref{fig1} (b), we conducted CKA~\cite{cortes2012algorithms} analysis on the feature spaces of three models in different scales (YOLOv7-X, L, and Tiny). It can be observed that larger scale differences lead to less similarity in model features. To this end, we aim to alleviate this issue during distillation.

To alleviate the diverse environments impact, existing object detection approaches typically utilize domain adaptation~\cite{kim2019self,yao2021multi,chen2020harmonizing} or domain generalization~\cite{lin2021domain,zhang2022towards,wu2022single} methods. However, domain adaptation approaches may not be readily employed when it is hard to guarantee the accessibility of the target domains. Most existing domain generalization method focus on disentangling object features from global features through convolution modules. However, the domain variations in UAV-OD tasks are more diverse than in general detection tasks.  Convolutional modules may not fully explore global semantic information~\cite{dosovitskiy2020image}, making it difficult to extract domain-invariant features. To alleviate this problem, we aim to extract and transfer global domain-invariant knowledge to the student.
% We seek to create a method to extract and transfer global domain-invariant knowledge to the student model.

In this paper, we propose a novel knowledge distillation framework for UAV-based Object Detection. 
% we employ CKA to analyze the differences in feature space similarity among detectors of different scales. The results indicate that models with similar scales have more similar features, making it easier to transfer information during the knowledge distillation process. 
Specially, considering that knowledge transfer is easier between models of similar scales, we design a progressive knowledge distillation approach by introducing an additional junior teacher model, as shown in Fig~\ref{fig1} (c). Then we utilize Fast Fourier Transform (FFT)~\cite{nussbaumer1982fast} to develop a novel knowledge transfer method to extract domain-invariant feature during distillation. It can mitigate domain shift issues caused by complex backgrounds in UAV-OD. Finally, through progressive teaching and the transfer of domain-invariant feature knowledge, the accuracy of the lightweight student model is effectively improved.

Our main contributions are highlighted as follows:
\begin{itemize}

\item We provide a progressive knowledge distillation framework for UAV-OD, which can also be applied to any distillation methods on object detection. %It can reduce the risk of information loss due to the significant difference in scale between the teacher and student models.

\item We propose a new feature transfer method that utilizes Fast Fourier Transform to extract key objects features for knowledge transfering, effectively enhancing the accuracy of the student model. %To the best of our knowledge, this is the first approach to introduce FFT in the knowledge distillation process of UAV-OD tasks.

\item We conduct extensive experiments on two UAV-OD datasets to verify our method. The results demonstrate that our method achieved SoTA performance. %It can also conclude that our distillation approach effectively excavate the potential of the student model, potentially surpassing the teacher model in accuracy.

\end{itemize}

\section{Method}
\begin{figure}[t]
    \centering
    \includegraphics[width=0.99\linewidth]{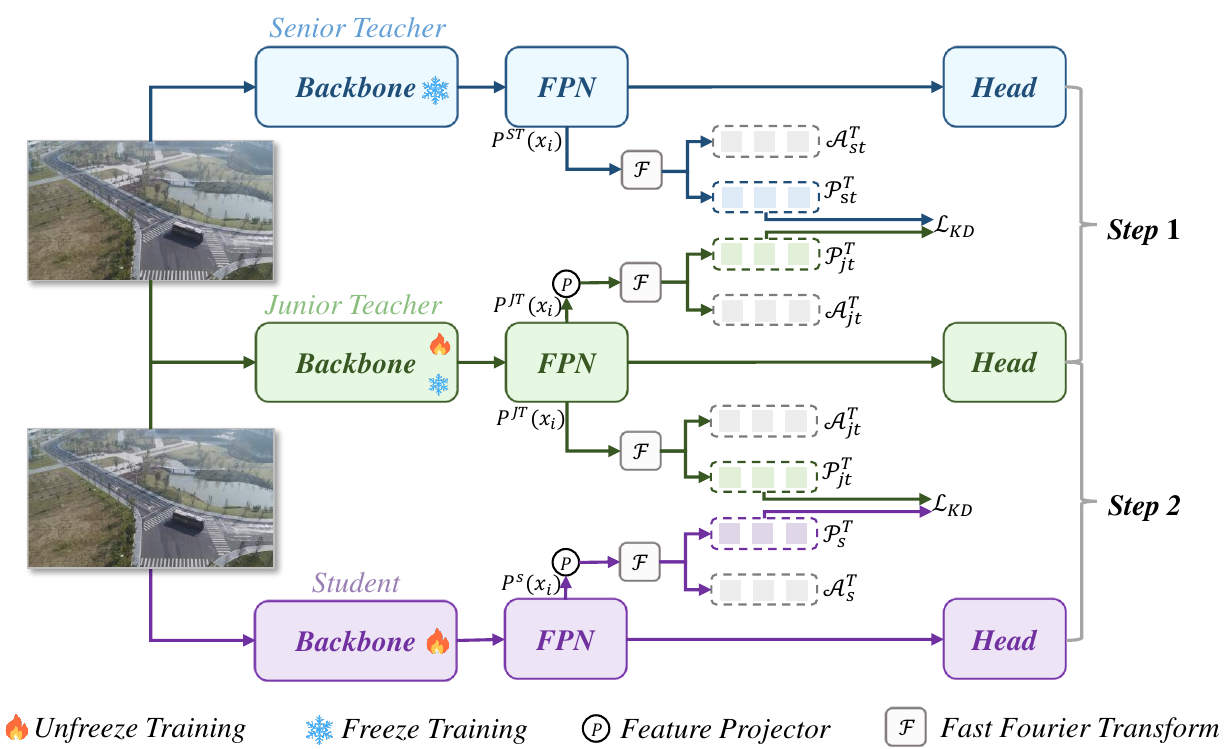}
    % \vspace{-0.3cm}
    \caption{Overview of our proposed method. We leverage two novel distillation methods for the UAV-OD task. \textbf{Step 1}: We introduce two different scale teacher models (senior teacher and junior teacher) and adopt a progressive distillation strategy. \textbf{Step 2}: We propose a domain-invariant knowledge transferring approach by utilizing fast fourier transform. Our approach can effectively excavate the potential of the student model in UAV-OD, enabling it to reach or exceed the junior teacher model's accuracy.}
    \label{fig2}
    % \vspace{-0.3cm}
\end{figure}

In this section, we introduce our proposed knowledge distillation framework. The overview is represented in Fig~\ref{fig2}. 
%Firstly, we revisit general knowledge distillation methods. Next, we present the progressive KD approach, which can excavate the potential of the student model. Finally, we provide a domain-invariant knowledge transfer strategy for the UAV-OD task, which can further enhance the efficiency of distillation. 

\subsection{Revisit General KD Methods}

In feature-based distillation for object detection, knowledge transfer usually utilizes multi-scale features from the FPN networks of both the teacher and student models. 
%Since the student model's feature maps have fewer channels than the teacher model, a feature projector $g(\cdot)$ is needed to align the channel numbers of the teacher model's feature maps. 
A feature projector $g(\cdot)$ is employed to match the channel numbers of the student model's feature maps with those of the teacher model.
% align the channel numbers of the student model's featuremaps with the more channels in the teacher model.
Subsequently, a distance metric function $d_{T}$ is used to minimize the feature discrepancy between the teacher and student models, such as the MSE~\cite{marmolin1986subjective} or SSIM~\cite{de2022structural}. The process is represented as follows:

\begin{equation}
    \begin{split}
        \resizebox{0.9\hsize}{!}{$\underset{\theta}{\operatorname{argmin}} \mathcal{L}_{KD}(\theta)=\frac{1}{N} \sum_{i}d_{T}\left(P^{T}\left(x_{i}\right), g\left(P^{S}\left(x_{i}, \theta\right)\right)\right)$},
    \end{split}
\end{equation}
where $x_{i}$ is the input image from the train-set, $N$ denotes the number of images, $\theta$ is the parameters of the student model, $P^{T}(x)$ and $P^{S}(x)$ represent the feature maps of the teacher and student models, respectively.

\begin{table*}[t]
% \renewcommand\arraystretch{1.05}
% \captionsetup{font=small}
\caption{Comparison with SoTA object detection KD methods on VisDrone and SynDrone with YOLOv7-L as the teacher and YOLOv7-Tiny as the student model. Black bold font represents the highest result, blue font represents the second highest result.}
\centering
%\resizebox{\textwidth}{!}{
\begin{tabular}{c|l|c|p{1.2cm}p{1.2cm}p{1.2cm}p{1.2cm}p{1.2cm}p{1.2cm}}
\toprule
Dataset & Method & Publication & mAP  & $AP_{0.5}$  & $AP_{0.75}$ & $AP_S$ & $AP_M$  & $AP_L$  \\
\hline
\multirow{6}{*}{VisDrone} &\cellcolor[HTML]{E0E0E0}YOLOv7-L (T) & \cellcolor[HTML]{E0E0E0}CVPR2023 & \cellcolor[HTML]{E0E0E0}\blue{16.94}   & \cellcolor[HTML]{E0E0E0}30.26                & \cellcolor[HTML]{E0E0E0}\textbf{16.89}  & \cellcolor[HTML]{E0E0E0}8.17     & \cellcolor[HTML]{E0E0E0}\textbf{26.90}   & \cellcolor[HTML]{E0E0E0}\textbf{42.41}  \\
        & \cellcolor[HTML]{E0E0E0}YOLOv7-Tiny (S)        & \cellcolor[HTML]{E0E0E0}CVPR2023 & \cellcolor[HTML]{E0E0E0}11.62      & \cellcolor[HTML]{E0E0E0}21.95   & \cellcolor[HTML]{E0E0E0}11.25   & \cellcolor[HTML]{E0E0E0}4.71  & \cellcolor[HTML]{E0E0E0}18.39    & \cellcolor[HTML]{E0E0E0}32.60     \\
        & FitNets~\cite{fitnets}   & ICLR2015 & 12.87   & 24.71   & 12.34   & 5.21    & 20.62   & 34.77   \\
        & BCKD~\cite{BCKD} & ICCV2023  & 16.08   & \blue{30.71}   & 15.81   & \blue{8.81}    & 24.90   & 26.60   \\
        & CrossKD~\cite{crosskd} & CVPR2024     & 14.96   & 29.22   & 14.11   & 8.79    & 23.77   & 24.54   \\
        &  \cellcolor[HTML]{DAE8FC}Ours & \cellcolor[HTML]{DAE8FC}- &  \cellcolor[HTML]{DAE8FC}\textbf{17.07}   &  \cellcolor[HTML]{DAE8FC}\textbf{31.92}   &  \cellcolor[HTML]{DAE8FC}\blue{16.77}   &  \cellcolor[HTML]{DAE8FC}\textbf{9.56}    &  \cellcolor[HTML]{DAE8FC}\blue{25.90}   &  \cellcolor[HTML]{DAE8FC}\blue{38.98}   \\
        % &   Improvements              & \begin{tabular}[c]{@{}l@{}}+0.99\\    +0.13\end{tabular} & \begin{tabular}[c]{@{}l@{}}+1.21\\    +1.66\end{tabular} & \begin{tabular}[c]{@{}l@{}}+0.96\\    -0.12\end{tabular} & \begin{tabular}[c]{@{}l@{}}+0.75\\    +1.39\end{tabular} & \begin{tabular}[c]{@{}l@{}}+1.00\\    -1.00\end{tabular} & \begin{tabular}[c]{@{}l@{}}+4.21\\    -3.43\end{tabular} \\
\hline
\multirow{6}{*}{SynDrone} 
                          & \cellcolor[HTML]{E0E0E0}YOLOv7-L (T)& \cellcolor[HTML]{E0E0E0}CVPR2023            & \cellcolor[HTML]{E0E0E0}\blue{33.66}   & \cellcolor[HTML]{E0E0E0}\blue{64.01}   & \cellcolor[HTML]{E0E0E0}\blue{30.36}   & \cellcolor[HTML]{E0E0E0}18.44   & \cellcolor[HTML]{E0E0E0}\blue{42.19}   & \cellcolor[HTML]{E0E0E0}\textbf{40.97}   \\
                          & \cellcolor[HTML]{E0E0E0}YOLOv7-Tiny (S)  & \cellcolor[HTML]{E0E0E0}CVPR2023       & \cellcolor[HTML]{E0E0E0}30.21   & \cellcolor[HTML]{E0E0E0}58.41   & \cellcolor[HTML]{E0E0E0}27.78   & \cellcolor[HTML]{E0E0E0}15.66   & \cellcolor[HTML]{E0E0E0}37.88   & \cellcolor[HTML]{E0E0E0}36.68   \\
                          & FitNets~\cite{fitnets}  &ICLR2015               & 32.12   & 61.73   & 29.32   & 16.82   & 40.21   & \blue{39.38}   \\
                          & BCKD~\cite{BCKD}    & ICCV2023              & 32.66   & 63.12   & 29.95   & \blue{21.52}   & 40.52   & 31.93   \\
                          & CrossKD~\cite{crosskd} & CVPR2024              & 31.79   & 61.85   & 28.46   & 17.84   & 39.53   & 30.88   \\
                          & \cellcolor[HTML]{DAE8FC}Ours & \cellcolor[HTML]{DAE8FC}-
                          &\cellcolor[HTML]{DAE8FC}\textbf{35.12}   & \cellcolor[HTML]{DAE8FC}\textbf{65.09}   & \cellcolor[HTML]{DAE8FC}\textbf{33.30}   & \cellcolor[HTML]{DAE8FC}\textbf{22.27}   & \cellcolor[HTML]{DAE8FC}\textbf{43.08}   & \cellcolor[HTML]{DAE8FC}36.51   \\
                          % &                       &         &         &         &         &         &        \\
\bottomrule

\end{tabular}
\label{tab:comparison}
\end{table*}

\subsection{Progressive Teaching} \label{proKD}

When teaching a lightweight detector like YOLOv7-tiny, existing methods typically select a larger-scale and high-performance pre-trained model as the teacher model. The aim is to leverage its great representation capability to ensure student model with higher accuracy. However, when teacher and student models have a large scale disparity, there will be a huge capacity gap between them~\cite{furlanello2018born,ren2023tinymim}, resulting in poor distillation results. 

% This capability gap can be likened to the real-life education environment. Generally, a doctoral student has profound knowledge in their specialized field, rich practical experience, and a systematic understanding of academic theory. Meanwhile, an undergraduate student has grasped some basic professional skills and theoretical knowledge. For a high school student, they are just beginning to understand some foundational subjects. It would be challenging for a computer science doctoral student to directly teach a high school student relevant knowledge because of the high school student's limited knowledge base and comprehension abilities. 
% However, if an undergraduate student is utilized as a bridge, allowing the undergraduate student to be guided by the doctoral student's advanced knowledge and feedback in teaching the high school student, more significant achievements can be made.

Therefore, we provide a progressive teaching strategy to alleviate the capacity gap between teacher and student models. We introduce two different scale teacher models (senior teacher and junior teacher) and adopt a two-stage distillation strategy to improve student model's accuracy. 

Utilizing YOLOv7 model as an example, we select YOLOv7-X as the senior teacher, YOLOv7-L as the junior teacher, and YOLOv7-Tiny as the student model. As represented in Fig~\ref{fig2}, in the first stage, the pre-trained senior teacher model (YOLOv7-X) is frozen, while the junior teacher model (YOLOv7-L) is unfrozen and trained. The junior teacher model will learn from the representation capabilities of the senior teacher to gain an advantage in guiding the student model. The process can be expressed as:
\begin{equation}
    \resizebox{0.9\hsize}{!}{$\underset{\theta}{\operatorname{argmin}} \mathcal{L}_{KD}(\theta_{JT})=\frac{1}{N} \sum_{i}d_{T}\left(P^{ST}\left(x_{i}\right), g\left(P^{JT}\left(x_{i}, \theta_{JT}\right)\right)\right)$},
\end{equation}
where $x_{i}$ is the input image from the train-set, $N$ denotes the number of images, $\theta_{JT}$ is the parameters of the junior teacher model, $P^{ST}(x)$ and $P^{JT}(x)$ represent the feature maps of the senior and junior teacher models, respectively. 

In the second stage, the junior teacher obtained from the previous stage is utilized to teach the student model. At this point, the parameters of the junior teacher are frozen, and only the parameters of the student model are updated through back-propagation. The training target function is as follows:
\begin{equation}
    \resizebox{0.9\hsize}{!}{$\underset{\theta}{\operatorname{argmin}} \mathcal{L}_{KD}(\theta_{S})=\frac{1}{N} \sum_{i}d_{T}\left(P^{JT}\left(x_{i}\right), g\left(P^{S}\left(x_{i}, \theta_{S}\right)\right)\right)$ ,}
\end{equation}
where $\theta_{S}$ is the parameters of the student model, the meanings of other symbols are consistent with the first stage.

\subsection{Domain-invariant Knowledge Transferring}
% As mentioned in section~\ref{proKD}, 
After progressive KD, student model can obtain stronger feature representation capabilities. However, UAV-OD task has a wide domain shift in images due to complex flying environments. It is still a challenge for the student model with poor domain-invariant feature extraction capabilities. Inspired by the spectral theorem that the frequency domain obeys the nature of global modeling~\cite{tang2024direct,wang2023generalized}, we propose to improve the domain generalization ability of student models via Fast Fourier Transform (FFT)~\cite{nussbaumer1982fast}:
\begin{equation}
    \mathcal{F}\left ( x \right ) \left ( u,v \right ) =\sum_{H-1}^{h=0} \sum_{W-1}^{w=0} x(h,w)e^{-j2\pi \left (  \frac{h}{H}u+\frac{w}{W}v \right ) }, 
\end{equation}
where $\mathcal{F}\left ( x \right )$ can be further decomposed to an amplitude spectrum $\mathcal{A} \left (  x\right )$ and a phase spectrum $\mathcal{P} \left (  x\right )$:
\begin{equation}
\begin{split}
      \mathcal{A} \left (  x\right ) \left (  u,v\right ) =\sqrt{\mathcal{R} ^{2}\left (  x\right ) \left ( u,v \right )+ \mathcal{I} ^{2}\left (  x\right ) \left ( u,v \right ) }, \\
     \mathcal{P} \left (  x\right ) \left (  u,v\right )=\arctan \left [  \frac{\mathcal{I}\left ( x \right ) \left ( u,v \right )  }{\mathcal{R}\left ( x \right ) \left ( u,v \right )  } \right ] ,
\end{split}
\end{equation}
where, $\mathcal{I}\left ( x \right )$ and $\mathcal{R}\left ( x \right )$ are the real and imaginary parts of $\mathcal{F}\left ( x \right )$. 

%Recent researchers conduct utilizing Fourier transformation to disentangle features in Deep Neural Networks (DNNs) into amplitude and phase, and independently processing these components to enhance model capabilities~\cite{lee2023decompose,chen2021amplitude}. 
The amplitude contains low-level features like color and brightness, which often show domain discrepancies, while the phase pertains to high-level semantic information that remains consistent across domains~\cite{xu2021fourier,lee2023decompose}.
% The amplitude includes low-level information such as color and brightness, which often exhibit domain discrepancies. And the phase is related to high-level semantic information that remain consistent across domains~\cite{xu2021fourier,lee2023decompose}.
% The limited feature extraction performance of the student model may be hindered by the complex cross-domain features in UAV images, affecting its training process. 
Therefore, we utilize FFT to separate the amplitude $\mathcal{A} \left (  x\right )$ and phase $\mathcal{P} \left (  x\right )$ in features, rendering them as domain-specific features and domain-invariant features, respectively. By employing domain-invariant features to teach the student model, we aim to further improve its accuracy and enhance distillation efficiency.

Assuming that the knowledge of the models, $(\mathrm{\textbf{x}}, F^{T})$, follows the distribution $\mathcal{U}$ on $\mathcal{X} \times \mathcal{Y}$, where $\mathcal{X}$ is the instance space of input images and $\mathcal{Y}$ is the fourier domain space of feature. Then our domain-invariant KD process can be presented as follows:
\begin{equation}
    \begin{split}
        \resizebox{0.9\hsize}{!}{$\underset{\theta}{\operatorname{argmin}} \hat{\mathcal{L}}_{KD}(\theta)=\mathbb{E}_{\mathcal{U}}\left\{d_{T}\left(\mathcal{F}\left(P^{T}\left(\mathrm{\textbf{x}}\right)\right), \mathcal{F}\left(g\left(P^{S}\left(\mathrm{\textbf{x}}, \theta\right)\right)\right)\right)\right\}$},
    \end{split}
    \label{formula6}
\end{equation}
where $\mathcal{F}$ denotes the FFT, $d_{T}$ denotes the distance metric function, $\theta$ is the parameters of the student model, $g(\cdot)$ is the feature projector, $P^{T}(x)$ and $P^{S}(x)$ represent the feature maps of the teacher and student models, respectively. 

We introduce the Equation~\ref{formula6} into the progressive distillation process in Section~\ref{proKD}, applying this process in both two stages. By minimizing the objective function in Equation~\ref{formula6}, we can transfer the domain-invariant knowledge of teacher model into the optimization process of the student model. 

\section{Experiments}

\subsection{Datasets and Evaluation Metrics}
To verify the effectiveness of our proposed approach, we adopted VisDrone~\cite{zhu2018vision} and SynDrone~\cite{rizzoli2023syndrone}, for experiments. VisDrone consists of 8629 aerial images belonging to 10 categories. Following previous UAV-OD work~\cite{CEASE,querydet}, we utilized 7019 images for training and 1610 images for testing. SynDrone is a synthetic large-scale multi-modal UAV-OD dataset, which consists of 60k aerial images in three different view angles. Due to SynDrone contains many similar images, we extracted 24,000 images from it, utilizing 21,600 images for training and 2,400 images for testing. We employed the mean Average Precision (mAP), Average Precision (AP)~\cite{everingham2010pascal} as the evaluation metrics on accuracy.

\subsection{Experimental Setup}
\subsubsection{Baselines} 
We adopted YOLOv7~\cite{wang2023yolov7} as the baseline models. Specially, we select YOLOv7-X as the senior teacher, YOLOv7-L as the junior teacher, YOLOv7-Tiny as the student model.  It's worth mentioning that in the \textbf{VisDrone Challenge of ICCV2023} (the authoritative challenge in UAV-OD), the champion team utilized YOLOv7 as their framework. Therefore, the experiments in this paper primarily focus on the YOLOv7 model. 

\subsubsection{Training Settings} All experiments were conducted in Pytorch with a NVIDIA RTX 3090 GPU. In first stage of progressive KD, we set the batch size to 16 and trained for 100 epochs. In other experiments, we trained for 100 epochs with batch size of 32.  All models were trained using an Adam optimizer~\cite{kingma2014adam} with a momentum of 0.937. The learning rate was initialized as 0.001 with a cosine decay~\cite{loshchilovstochastic}. 

\begin{table}[]
\centering
\caption{Ablation on Progressive Knowledge Distillation.}
\begin{tabular}{ccccc}
\toprule
Dataset                   & Teacher           & Student                      & $AP_{0.5}$ & $AP_{0.75}$ \\
\hline
\multirow{4}{*}{VisDrone} & -                 & \multirow{4}{*}{YOLOv7-Tiny} & 21.95  & 11.25   \\
                          & YOLOv7-L          &                              & 24.68  & 12.34   \\
                          & YOLOv7-X          &                              & 25.21  & 12.37   \\
                          & \cellcolor[HTML]{DAE8FC}YOLOv7-X→L &                              & \cellcolor[HTML]{DAE8FC}\textbf{26.36}  & \cellcolor[HTML]{DAE8FC}\textbf{13.01}   \\
\hline                    
\multirow{4}{*}{SynDrone} & -                 & \multirow{4}{*}{YOLOv7-Tiny} & 58.41  & 27.78   \\
                          & YOLOv7-L          &                              & 61.73  & 29.32   \\
                          & YOLOv7-X          &                              & 61.86  & 29.95   \\
                          & \cellcolor[HTML]{DAE8FC}YOLOv7-X→L &                              & \cellcolor[HTML]{DAE8FC}\textbf{62.36}  & \cellcolor[HTML]{DAE8FC}\textbf{29.99}  \\
\bottomrule                          
\end{tabular}
\label{tab:progressive}
\end{table}

\subsection{Experimental Results and Analyses}
% To explore the superiority of our approach, we conduct comparative experiments with the previous State-of-the-art (SoTA) KD methods on YOLOv7. As demonstrated in Table~\ref{tab:comparison}, the experimental results demonstrate that our method achieved SoTA performance across two UAV-OD datasets. Specifically, our method achieved an mAP of 17.07\% on the VisDrone dataset, surpassing the previous state-of-the-art method BCKD (16.08\% mAP) by 0.99\%. Furthermore, it surpasses the teacher model by 0.13\%. Our approach outperforms other knowledge distillation methods on all 6 metrics, with 3 metrics surpassing the teacher model. On the SynDrone dataset, our method achieved a mAP of 35.12\%, surpassing BCKD (32.66\% mAP) by 2.46\% and exceeding the teacher model (33.66\% mAP) by 1.46\%. Our method outperforms other KD approaches and the teacher model, with 5 metrics showing superior performance. These results above confirm the superiority of our method. It is worth mentioning that the results surpassing the teacher model demonstrate our method's ability to significantly unleash the potential of student models, thereby greatly enhancing the accuracy of student models on the UAV-OD task.
To explore the superiority of our approach, we conducted comparative experiments with previous SoTA KD methods on YOLOv7. As shown in Table~\ref{tab:comparison}, our method achieved SoTA performance on two UAV-OD datasets. On the VisDrone dataset, we achieved an mAP of 17.07\%, surpassing the previous SoTA method BCKD (16.08\%) by 0.99\% and exceeding the teacher model by 0.13\%. Our approach outperformed other distillation methods across all 6 metrics, with 3 metrics exceeding the teacher model. On the SynDrone dataset, we achieved an mAP of 35.12\%, exceeding BCKD (32.66\%) by 2.46\% and the teacher model (33.66\%) by 1.46\%. Overall, our method demonstrates superior performance on all metrics and effectively unleashes the potential of student models, significantly enhancing accuracy in UAV-OD tasks.
 
\begin{table}[]
\centering
\caption{Effectiveness analysis on our proposed progressive and domain-invariant KD.}
\begin{tabular}{
>{\centering\arraybackslash}p{1.2cm}
>{\centering\arraybackslash}p{1cm}
>{\centering\arraybackslash}p{1cm}
>{\centering\arraybackslash}p{1cm}
>{\centering\arraybackslash}p{1cm}}
\toprule
Dataset  & Progressive & FFT & $AP_{0.5}$ & $AP_{0.75}$ \\
\hline
\multirow{4}{*}{VisDrone} &      &          & 21.95  & 11.25   \\
                          &      & \checkmark        & 30.71  & 15.81   \\
                          & \checkmark    &          & 26.36  & 13.01   \\
                          & \cellcolor[HTML]{DAE8FC}\checkmark    & \cellcolor[HTML]{DAE8FC}\checkmark        & \cellcolor[HTML]{DAE8FC}\textbf{31.55}  & \cellcolor[HTML]{DAE8FC}\textbf{16.77}   \\
\hline
\multirow{4}{*}{SynDrone} &      &          & 58.41  & 27.78   \\
                          &      & \checkmark        & 62.50  & 30.76   \\
                          & \checkmark    &          & 62.36  & 29.99   \\
                          & \cellcolor[HTML]{DAE8FC}\checkmark    & \cellcolor[HTML]{DAE8FC}\checkmark     & \cellcolor[HTML]{DAE8FC}\textbf{64.71} & \cellcolor[HTML]{DAE8FC}\textbf{31.38}  \\
\bottomrule
\end{tabular}
\label{tab3}
\end{table}

\subsection{Ablation Studies}
To investigate the effectiveness of our approach, we conducted extensive ablation experiments.  It includes an effectiveness analysis of the two distillation strategies, an ablation study of progressive distillation, and the impact of the position of the domain-invariant feature distillation module.

\subsubsection{Ablation on Progressive KD} 
% As illustrated in Table~\ref{tab:progressive}, we conducted ablation study to validate the effectiveness of our progressive KD. Specifically, we compared the results obtained by directly utilizing YOLOv7-X and YOLOv7-L as teacher models for YOLOv7-Tiny distillation with the results from our progressive KD approach. 
% While distillation can significantly improve the performance of the student model, the model accuracy after being taught by YOLOv7-X is similar to that after being taught by YOLOv7-L. For example, only 0.53\% and 0.13\% higher on $AP_{0.5}$ on VisDrone and SynDrone, respectively. With our progressive distillation strategy, the accuracy of the student model can be further enhanced. For instance, on the VisDrone dataset, the $AP_{0.5}$ is improved by 1.15\% compared to employing YOLOv7-X as the teacher model. This result occurs because there is a significant difference in parameter size between YOLOv7-X and YOLOv7-Tiny, leading to a substantial disparity in knowledge space. Direct teaching is unable to fully leverage the accuracy advantage of the YOLOv7-X model due to the large knowledge gap. 
We conducted an ablation study to validate the effectiveness of our progressive knowledge distillation (KD), as illustrated in Table~\ref{tab:progressive}. We compared the performance of directly using YOLOv7-X and YOLOv7-L as teacher models for distilling YOLOv7-Tiny with our progressive KD approach. %Although distillation significantly improves the student model's performance, 
The accuracy improvements after distillation with YOLOv7-X is similar to that with YOLOv7-L, with only 0.53\% and 0.13\% increases in $AP_{0.5}$ on VisDrone and SynDrone, respectively. In contrast, our progressive distillation strategy further enhances accuracy; for instance, on the VisDrone dataset, $AP_{0.5}$ improves by 1.15\% compared to using YOLOv7-X as the teacher model. This result occurs due to the substantial parameter size difference between YOLOv7-X and YOLOv7-Tiny, which creates a significant knowledge gap, limiting the benefits of direct teaching.

\subsubsection{Effectiveness of two KD strategies} 
% We investigated the effectiveness of our progressive KD and domain-invariant KD approaches. The experimental results are demonstrated in Table~\ref{tab3}. By employing progressive KD, the $AP_{0.5}$ on VisDrone and SynDrone increased by 4.41\% and 3.95\%, respectively. By adopting domain-invariant KD, the student model's detection accuracy ($AP_{0.5}$) grows by 8.76\% and 4.09\% on two datasets, respectively. When both strategies are utilized together, the model's accuracy shows further improvement compared to using each method individually. The results indicates that both progressive KD and domain-invariant KD are effective and contribute to the final performance improvements. 
We investigated the effectiveness of our progressive KD and domain-invariant KD approaches, as shown in Table~\ref{tab3}. Progressive KD improved $AP_{0.5}$ by 4.41\% on VisDrone and 3.95\% on SynDrone. Domain-invariant KD increased detection accuracy by 8.76\% and 4.09\% on the two datasets, respectively. When combined, these strategies offer further accuracy improvements over utilizing each method individually, demonstrating their effectiveness in enhancing overall performance.

% 1.76\% 

\begin{table}[]
\centering
\caption{Influence of fast fourier transform utilization on VisDrone. `P3-P5' represents different layers of FPN.}
\begin{tabular}{>{\centering\arraybackslash}p{0.9cm}>{\centering\arraybackslash}p{0.9cm}>{\centering\arraybackslash}p{0.9cm}>{\centering\arraybackslash}p{0.9cm}>{\centering\arraybackslash}p{0.9cm}}
\toprule
P3 & P4 & P5 & $AP_{0.5}$ & $AP_{0.75}$ \\
\hline
\checkmark  &    &    & 29.37  & 14.07   \\
   & \checkmark  &    & 29.19  & 13.73   \\
   &    & \checkmark  & 28.91  & 13.82   \\
\hline
\checkmark  & \checkmark  &    & 31.15  & 15.19   \\
   & \checkmark  & \checkmark  & 30.38  & 15.33   \\
\checkmark  &    & \checkmark  & 29.99  & 15.08   \\
\hline
\cellcolor[HTML]{DAE8FC}\checkmark  & \cellcolor[HTML]{DAE8FC}\checkmark  & \cellcolor[HTML]{DAE8FC}\checkmark  & \cellcolor[HTML]{DAE8FC}\textbf{31.55}  & \cellcolor[HTML]{DAE8FC}\textbf{16.77}  \\
\bottomrule
\end{tabular}
\label{tab4}
\end{table}

\begin{figure}[]
    \centering
    \includegraphics[width=0.99\linewidth]{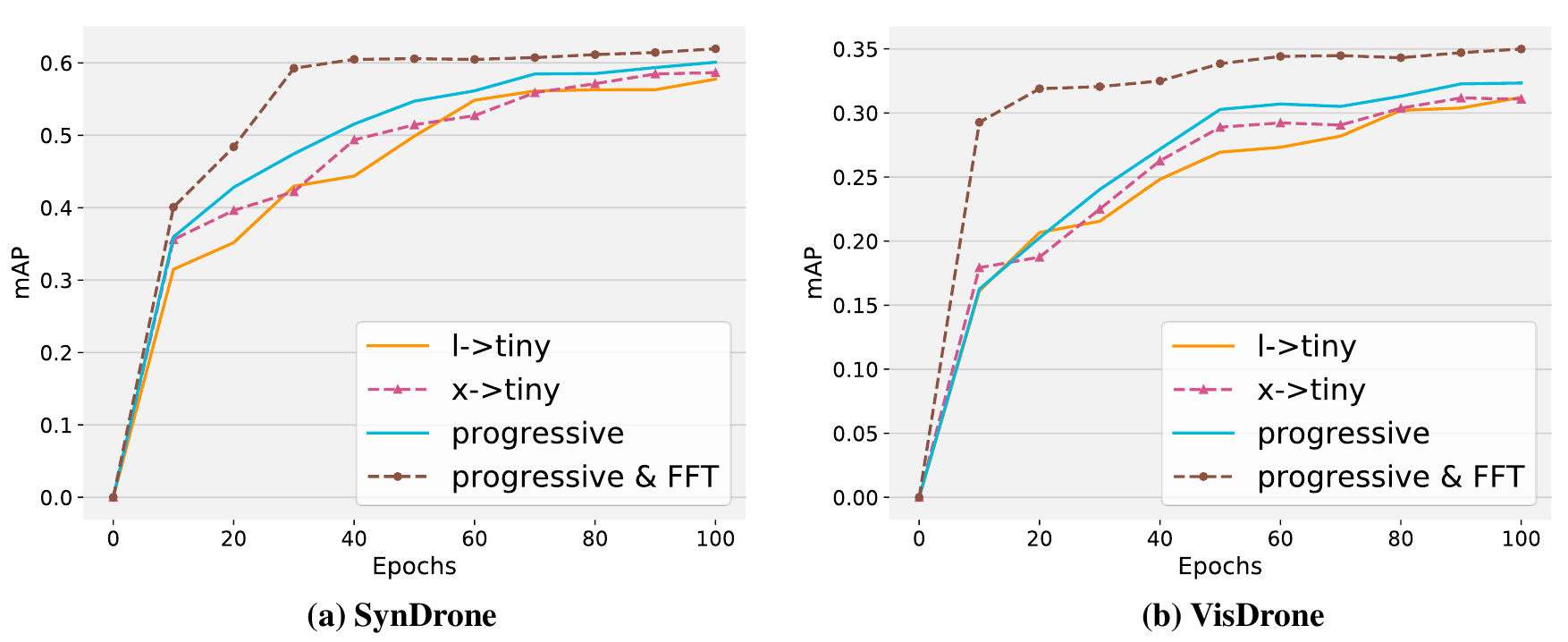}
    % \vspace{-0.3cm}
    \caption{mAP curves during distillation on SynDrone and VisDrone Datasets.}
    \label{fig3}
    \vspace{-0.3cm}
\end{figure}

\subsubsection{FFT on Different FPN layers}
% We compare different locations and numbers of the domain-invariant feature extraction modules on FPN layers and present the reults in Table~\ref{tab4}. From the table, we can observe that the detection accuracy gradually increases as the number of feature extraction modules increases, reaching its peak in the three. The result underscores the importance of transferring domain-invariant feature knowledge across all parts of the FPN.
% It can also be noted that when there is only one module, adding it to the P3 layer provides the most significant improvement in model accuracy. This is because in UAV-OD task, there are large number of small objects. The P3 layer, as a high-resolution feature layer, is primarily responsible for detecting small-scale objects. Therefore, transferring domain-invariant knowledge in this layer is more critical.
We compared various placements and quantities of domain-invariant feature extraction modules on FPN layers, presented in Table~\ref{tab4}. The results indicate that detection accuracy improves with the number of modules, peaking at three. This highlights the importance of transferring domain-invariant feature knowledge throughout the FPN. Additionally, with only one module, placing it in the P3 layer yields the most significant accuracy improvement due to its role in detecting small objects, crucial in UAV-OD tasks. Therefore, transferring domain-invariant knowledge in this high-resolution layer is essential.

\subsubsection{Ablation on Distillation Efficiency}
We plotted the mAP curves for different distillation strategies, as shown in Fig~\ref{fig3}. From the trend of the curves, it can be observed that after the addition of domain-invariant knowledge transfer, the model shows significant improvements in both accuracy and convergence speed. It is because the student model learns directly from domain-invariant object features during the distillation process, while disregarding domain-specific features.
%It is evident that our domain-invariant feature KD strategy not only improves the accuracy of the student model but also accelerates the model's convergence speed. It is because the student model learns the object features more efficiently.

\section{Conclusion}
In this paper, we introduced a novel knowledge distillation framework for UAV-based Object Detection to enhance the lightweight student model's accuracy. Specially, we proposed a progressive KD method to alleviate knowledge gap caused by scale disparities. Then we designed a domain-invariant knowledge transfer approach to deal with the situation of domain variations in UAV-OD. Furthermore, we conducted extensive experiments to demonstrate the superiority of our approach. The results indicates that our distillation approach effectively excavate the potential of the student model, potentially surpassing the teacher model in accuracy.
\bibliographystyle{IEEEtran}
\bibliography{sample-base}

\begin{thebibliography}{10}
\providecommand{\url}[1]{#1}
\csname url@samestyle\endcsname
\providecommand{\newblock}{\relax}
\providecommand{\bibinfo}[2]{#2}
\providecommand{\BIBentrySTDinterwordspacing}{\spaceskip=0pt\relax}
\providecommand{\BIBentryALTinterwordstretchfactor}{4}
\providecommand{\BIBentryALTinterwordspacing}{\spaceskip=\fontdimen2\font plus
\BIBentryALTinterwordstretchfactor\fontdimen3\font minus \fontdimen4\font\relax}
\providecommand{\BIBforeignlanguage}[2]{{%
\expandafter\ifx\csname l@#1\endcsname\relax
\typeout{** WARNING: IEEEtran.bst: No hyphenation pattern has been}%
\typeout{** loaded for the language `#1'. Using the pattern for}%
\typeout{** the default language instead.}%
\else
\language=\csname l@#1\endcsname
\fi
#2}}
\providecommand{\BIBdecl}{\relax}
\BIBdecl

\bibitem{huang2022object}
H.~Huang, Y.~Tang, Z.~Tan, J.~Zhuang, C.~Hou, W.~Chen, and J.~Ren, ``Object-based attention mechanism for color calibration of uav remote sensing images in precision agriculture,'' \emph{IEEE transactions on geoscience and remote sensing}, vol.~60, pp. 1--13, 2022.

\bibitem{luo2023evolutionary}
S.~Luo, H.~Li, Y.~Li, C.~Shao, H.~Shen, and L.~Zhang, ``An evolutionary shadow correction network and a benchmark uav dataset for remote sensing images,'' \emph{IEEE Transactions on Geoscience and Remote Sensing}, 2023.

\bibitem{han2022comprehensive}
Y.~Han, H.~Liu, Y.~Wang, and C.~Liu, ``A comprehensive review for typical applications based upon unmanned aerial vehicle platform,'' \emph{IEEE Journal of Selected Topics in Applied Earth Observations and Remote Sensing}, vol.~15, pp. 9654--9666, 2022.

\bibitem{lopez2022unmanned}
Y.~{\'A}. L{\'o}pez, M.~Garcia-Fernandez, G.~Alvarez-Narciandi, and F.~L.-H. Andr{\'e}s, ``Unmanned aerial vehicle-based ground-penetrating radar systems: A review,'' \emph{IEEE Geoscience and Remote Sensing Magazine}, vol.~10, no.~2, pp. 66--86, 2022.

\bibitem{zhang2024empowering}
Y.~Zhang, Z.~Gong, W.~Liu, H.~Wen, P.~Wan, J.~Qi, X.~Hu, and P.~Zhong, ``Empowering physical attacks with jacobian matrix regularization against vit-based detectors in uav remote sensing images,'' \emph{IEEE Transactions on Geoscience and Remote Sensing}, 2024.

\bibitem{wu2021deep}
X.~Wu, W.~Li, D.~Hong, R.~Tao, and Q.~Du, ``Deep learning for unmanned aerial vehicle-based object detection and tracking: A survey,'' \emph{IEEE Geoscience and Remote Sensing Magazine}, vol.~10, no.~1, pp. 91--124, 2021.

\bibitem{zitar2023review}
R.~A. Zitar, M.~Al-Betar, M.~Ryalat, and S.~Kassaymehd, ``A review of {UAV} visual detection and tracking methods,'' \emph{arXiv preprint arXiv:2306.05089}, 2023.

\bibitem{han2015learning}
S.~Han, J.~Pool, J.~Tran, and W.~Dally, ``Learning both weights and connections for efficient neural network,'' \emph{Advances in neural information processing systems}, vol.~28, 2015.

\bibitem{fang2023depgraph}
G.~Fang, X.~Ma, M.~Song, M.~B. Mi, and X.~Wang, ``Depgraph: Towards any structural pruning,'' in \emph{Proceedings of the IEEE/CVF conference on computer vision and pattern recognition}, 2023, pp. 16\,091--16\,101.

\bibitem{han2015deep}
S.~Han, H.~Mao, and W.~J. Dally, ``Deep compression: Compressing deep neural networks with pruning, trained quantization and huffman coding,'' \emph{arXiv preprint arXiv:1510.00149}, 2015.

\bibitem{hinton2015distilling}
G.~Hinton, O.~Vinyals, and J.~Dean, ``Distilling the knowledge in a neural network,'' \emph{arXiv preprint arXiv:1503.02531}, 2015.

\bibitem{li2023object}
Z.~Li, P.~Xu, X.~Chang, L.~Yang, Y.~Zhang, L.~Yao, and X.~Chen, ``When object detection meets knowledge distillation: A survey,'' \emph{IEEE Transactions on Pattern Analysis and Machine Intelligence}, vol.~45, no.~8, pp. 10\,555--10\,579, 2023.

\bibitem{cao2022pkd}
W.~Cao, Y.~Zhang, J.~Gao, A.~Cheng, K.~Cheng, and J.~Cheng, ``Pkd: General distillation framework for object detectors via pearson correlation coefficient,'' \emph{Advances in Neural Information Processing Systems}, vol.~35, pp. 15\,394--15\,406, 2022.

\bibitem{yang2022focal}
Z.~Yang, Z.~Li, X.~Jiang, Y.~Gong, Z.~Yuan, D.~Zhao, and C.~Yuan, ``Focal and global knowledge distillation for detectors,'' in \emph{Proceedings of the IEEE/CVF Conference on Computer Vision and Pattern Recognition}, 2022, pp. 4643--4652.

\bibitem{zhang2023structured}
L.~Zhang and K.~Ma, ``Structured knowledge distillation for accurate and efficient object detection,'' \emph{IEEE Transactions on Pattern Analysis and Machine Intelligence}, 2023.

\bibitem{wang2024crosskd}
J.~Wang, Y.~Chen, Z.~Zheng, X.~Li, M.-M. Cheng, and Q.~Hou, ``Crosskd: Cross-head knowledge distillation for object detection,'' in \emph{Proceedings of the IEEE/CVF Conference on Computer Vision and Pattern Recognition}, 2024, pp. 16\,520--16\,530.

\bibitem{cortes2012algorithms}
C.~Cortes, M.~Mohri, and A.~Rostamizadeh, ``Algorithms for learning kernels based on centered alignment,'' \emph{The Journal of Machine Learning Research}, vol.~13, pp. 795--828, 2012.

\bibitem{kim2019self}
S.~Kim, J.~Choi, T.~Kim, and C.~Kim, ``Self-training and adversarial background regularization for unsupervised domain adaptive one-stage object detection,'' in \emph{Proceedings of the IEEE/CVF International Conference on Computer Vision}, 2019, pp. 6092--6101.

\bibitem{yao2021multi}
X.~Yao, S.~Zhao, P.~Xu, and J.~Yang, ``Multi-source domain adaptation for object detection,'' in \emph{Proceedings of the IEEE/CVF International Conference on Computer Vision}, 2021, pp. 3273--3282.

\bibitem{chen2020harmonizing}
C.~Chen, Z.~Zheng, X.~Ding, Y.~Huang, and Q.~Dou, ``Harmonizing transferability and discriminability for adapting object detectors,'' in \emph{Proceedings of the IEEE/CVF conference on computer vision and pattern recognition}, 2020, pp. 8869--8878.

\bibitem{lin2021domain}
C.~Lin, Z.~Yuan, S.~Zhao, P.~Sun, C.~Wang, and J.~Cai, ``Domain-invariant disentangled network for generalizable object detection,'' in \emph{Proceedings of the IEEE/CVF international conference on computer vision}, 2021, pp. 8771--8780.

\bibitem{zhang2022towards}
X.~Zhang, Z.~Xu, R.~Xu, J.~Liu, P.~Cui, W.~Wan, C.~Sun, and C.~Li, ``Towards domain generalization in object detection,'' \emph{arXiv preprint arXiv:2203.14387}, 2022.

\bibitem{wu2022single}
A.~Wu and C.~Deng, ``Single-domain generalized object detection in urban scene via cyclic-disentangled self-distillation,'' in \emph{Proceedings of the IEEE/CVF Conference on computer vision and pattern recognition}, 2022, pp. 847--856.

\bibitem{dosovitskiy2020image}
A.~Dosovitskiy, L.~Beyer, A.~Kolesnikov, D.~Weissenborn, X.~Zhai, T.~Unterthiner, M.~Dehghani, M.~Minderer, G.~Heigold, S.~Gelly \emph{et~al.}, ``An image is worth 16x16 words: Transformers for image recognition at scale,'' \emph{arXiv preprint arXiv:2010.11929}, 2020.

\bibitem{nussbaumer1982fast}
H.~J. Nussbaumer and H.~J. Nussbaumer, \emph{The fast Fourier transform}.\hskip 1em plus 0.5em minus 0.4em\relax Springer, 1982.

\bibitem{marmolin1986subjective}
H.~Marmolin, ``Subjective mse measures,'' \emph{IEEE transactions on systems, man, and cybernetics}, vol.~16, no.~3, pp. 486--489, 1986.

\bibitem{de2022structural}
P.~De~Rijk, L.~Schneider, M.~Cordts, and D.~Gavrila, ``Structural knowledge distillation for object detection,'' \emph{Advances in Neural Information Processing Systems}, vol.~35, pp. 3858--3870, 2022.

\bibitem{fitnets}
A.~Romero, N.~Ballas, S.~E. Kahou, A.~Chassang, C.~Gatta, and Y.~Bengio, ``Fitnets: Hints for thin deep nets,'' \emph{arXiv preprint arXiv:1412.6550}, 2014.

\bibitem{BCKD}
L.~Yang, X.~Zhou, X.~Li, L.~Qiao, Z.~Li, Z.~Yang, G.~Wang, and X.~Li, ``Bridging cross-task protocol inconsistency for distillation in dense object detection,'' in \emph{Proceedings of the IEEE/CVF International Conference on Computer Vision}, 2023, pp. 17\,175--17\,184.

\bibitem{crosskd}
J.~Wang, Y.~Chen, Z.~Zheng, X.~Li, M.-M. Cheng, and Q.~Hou, ``Crosskd: Cross-head knowledge distillation for object detection,'' in \emph{Proceedings of the IEEE/CVF Conference on Computer Vision and Pattern Recognition}, 2024, pp. 16\,520--16\,530.

\bibitem{furlanello2018born}
T.~Furlanello, Z.~Lipton, M.~Tschannen, L.~Itti, and A.~Anandkumar, ``Born again neural networks,'' in \emph{International conference on machine learning}.\hskip 1em plus 0.5em minus 0.4em\relax PMLR, 2018, pp. 1607--1616.

\bibitem{ren2023tinymim}
S.~Ren, F.~Wei, Z.~Zhang, and H.~Hu, ``Tinymim: An empirical study of distilling mim pre-trained models,'' in \emph{Proceedings of the IEEE/CVF Conference on Computer Vision and Pattern Recognition}, 2023, pp. 3687--3697.

\bibitem{tang2024direct}
J.~Tang, S.~Chen, G.~Niu, H.~Zhu, J.~T. Zhou, C.~Gong, and M.~Sugiyama, ``Direct distillation between different domains,'' \emph{arXiv preprint arXiv:2401.06826}, 2024.

\bibitem{wang2023generalized}
K.~Wang, X.~Fu, Y.~Huang, C.~Cao, G.~Shi, and Z.-J. Zha, ``Generalized uav object detection via frequency domain disentanglement,'' in \emph{Proceedings of the IEEE/CVF conference on computer vision and pattern recognition}, 2023, pp. 1064--1073.

\bibitem{xu2021fourier}
Q.~Xu, R.~Zhang, Y.~Zhang, Y.~Wang, and Q.~Tian, ``A fourier-based framework for domain generalization,'' in \emph{Proceedings of the IEEE/CVF conference on computer vision and pattern recognition}, 2021, pp. 14\,383--14\,392.

\bibitem{lee2023decompose}
S.~Lee, J.~Bae, and H.~Y. Kim, ``Decompose, adjust, compose: Effective normalization by playing with frequency for domain generalization,'' in \emph{Proceedings of the IEEE/CVF conference on computer vision and pattern recognition}, 2023, pp. 11\,776--11\,785.

\bibitem{zhu2018vision}
P.~Zhu, L.~Wen, X.~Bian, H.~Ling, and Q.~Hu, ``Vision meets drones: A challenge,'' \emph{arXiv preprint arXiv:1804.07437}, 2018.

\bibitem{rizzoli2023syndrone}
G.~Rizzoli, F.~Barbato, M.~Caligiuri, and P.~Zanuttigh, ``Syndrone-multi-modal {UAV} dataset for urban scenarios,'' in \emph{Proceedings of the IEEE/CVF International Conference on Computer Vision}, 2023.

\bibitem{CEASE}
B.~Du, Y.~Huang, J.~Chen, and D.~Huang, ``Adaptive sparse convolutional networks with global context enhancement for faster object detection on drone images,'' in \emph{Proceedings of the IEEE/CVF conference on computer vision and pattern recognition}, 2023, pp. 13\,435--13\,444.

\bibitem{querydet}
C.~Yang, Z.~Huang, and N.~Wang, ``Querydet: Cascaded sparse query for accelerating high-resolution small object detection,'' in \emph{Proceedings of the IEEE/CVF Conference on computer vision and pattern recognition}, 2022, pp. 13\,668--13\,677.

\bibitem{everingham2010pascal}
M.~Everingham, L.~Van~Gool, C.~K. Williams, J.~Winn, and A.~Zisserman, ``The {Pascal} visual object classes ({VOC}) challenge,'' \emph{International journal of computer vision}, 2010.

\bibitem{wang2023yolov7}
C.-Y. Wang, A.~Bochkovskiy, and H.-Y.~M. Liao, ``Yolov7: Trainable bag-of-freebies sets new state-of-the-art for real-time object detectors,'' in \emph{Proceedings of the IEEE/CVF Conference on Computer Vision and Pattern Recognition}, 2023, pp. 7464--7475.

\bibitem{kingma2014adam}
D.~P. Kingma and J.~Ba, ``Adam: A method for stochastic optimization,'' \emph{arXiv preprint arXiv:1412.6980}, 2014.

\bibitem{loshchilovstochastic}
I.~Loshchilov and F.~Hutter, ``Stochastic gradient descent with warm restarts,'' in \emph{International Conference on Learning Representations}, 2017, pp. 1--16.

\end{thebibliography}

\vfill

\end{document}